\documentclass[10pt,twocolumn,letterpaper]{article}

\usepackage{iccv}
\usepackage{times}
\usepackage{epsfig}
\usepackage{graphicx}
\usepackage{amsmath}
\usepackage{amssymb}

\usepackage{microtype}
\usepackage{subfigure}
\usepackage{booktabs} 
\usepackage{multirow}
\usepackage{caption}
\usepackage{bbm}
\usepackage{setspace}
\usepackage[ruled,vlined]{algorithm2e}

\usepackage{hyperref}

\def\eg{\emph{e.g}.} 
\def\ie{\emph{i.e}.}

\newcommand{\tabincell}[2]{\begin{tabular}{@{}#1@{}}#2\end{tabular}} 

\iccvfinalcopy 

\def\iccvPaperID{11228} 

\ificcvfinal\fi

\begin{document}

\title{Generative Modeling for Multi-task Visual Learning}

\author{
{Zhipeng Bao$^{1}$ \qquad Yu-Xiong Wang$^{2}$ \qquad Martial Hebert$^1$} \\
{$^1$Carnegie Mellon University \qquad $^2$University of Illinois at Urbana-Champaign}\\
  \texttt{\{zbao, hebert\}@cs.cmu.edu \qquad yxw@illinois.edu} \\
}

\maketitle
\ificcvfinal\thispagestyle{empty}\fi

\begin{abstract}
   Generative modeling has recently shown great promise in computer vision, but it has mostly focused on synthesizing visually realistic images. In this paper, motivated by multi-task learning of shareable feature representations, we consider a novel problem of learning a shared generative model that is useful across various visual perception tasks. Correspondingly, we propose a general multi-task oriented generative modeling (MGM) framework, by coupling a discriminative multi-task network with a generative network. While it is challenging to synthesize both RGB images and pixel-level annotations in multi-task scenarios, our framework enables us to use synthesized images paired with only weak annotations (\ie, image-level scene labels) to facilitate multiple visual tasks. Experimental evaluation on challenging multi-task benchmarks, including NYUv2 and Taskonomy, demonstrates that our MGM framework improves the performance of all the tasks by large margins, consistently outperforming state-of-the-art multi-task approaches.
\end{abstract}


\section{Introduction}
\label{sec:intro}
\begin{figure*}[t]
    \centering
    \includegraphics[width =  0.9\linewidth]{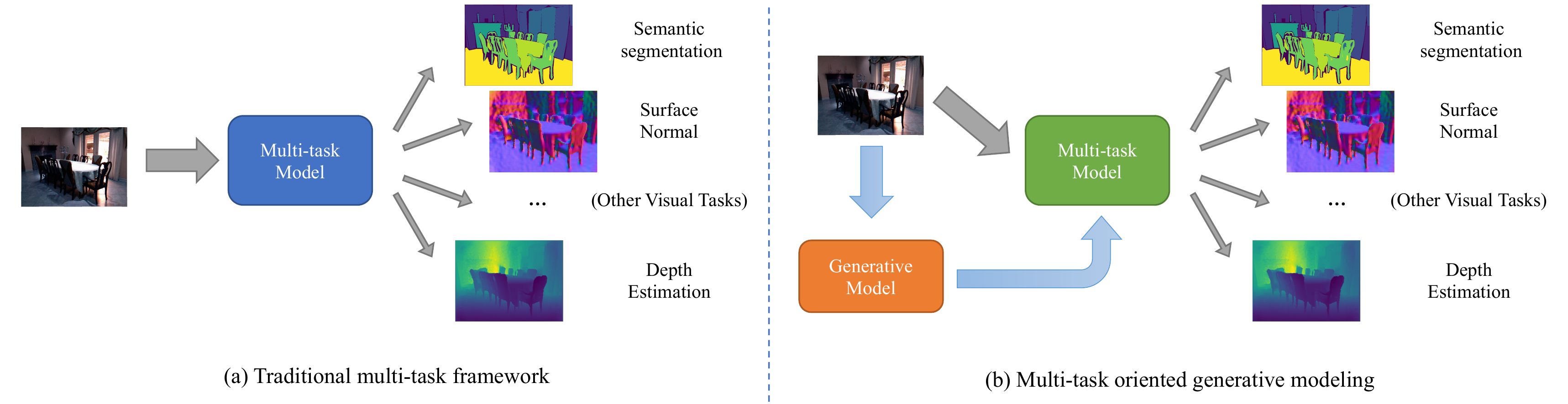}
    \vspace{-10pt}
    \caption{\textbf{Left:} Traditional multi-task learning framework (that learns shared feature representations) \textbf{vs.} \textbf{Right:} our proposed multi-task oriented generative modeling (that learns a shared generative model across various visual perception tasks)}
    \label{fig:teaser}
    \vspace{-16pt}
\end{figure*}
Seeing with the mind's eye --- creating internal images of objects and scenes not actually present to the senses, is perhaps one of the hallmarks in human cognition~\cite{pelaprat2011minding}. For humans, this visual imagination integrates learning experience and facilitates learning by solving different problems~\cite{egan2014imagination,pearson2019human,pelaprat2011minding,egan1989memory}. Inspired by such ability, there has been increasing interest in building generative models that can synthesize images~\cite{goodfellow2014generative}. Yet, most of the effort has focused on generating visually realistic images~\cite{brock2018large,zhang2019self}, which are still far from useful for machine perception tasks~\cite{shmelkov2018good,borji2019pros,wu2016quantitative}. Even though recent work has started improving the ``usefulness'' of synthesized images, this line of investigation is often limited to a single specific task~\cite{nguyen2018rendernet,sitzmann2019deepvoxels,zhu2018emotion,souly2017semi}. Could we guide generative models to benefit {\em multiple} visual tasks?

While similar spirits of shared feature representations have been widely studied as multi-task learning or meta-learning~\cite{finn2017model,zamir2018taskonomy}, here we are taking a different perspective --- {\em learning a shared generative model across various tasks} (as illustrated in Figure~\ref{fig:teaser}). Leveraging multiple tasks allows us to capture the underlying image generation mechanism for more comprehensive object and scene understanding than being done within individual tasks. Taking simultaneous semantic segmentation, depth estimation, and surface normal prediction as an example (Figure~\ref{fig:teaser}), successful generative modeling requires understanding not only the semantics but also the 3D geometric structure and physical property of the input image. Meanwhile, a learned common generative model facilitates the flow of knowledge across tasks, so that they benefit one another. For instance, the synthesized images provide meaningful variations in existing images and could be used as additional training data to build better task-specific models.

This paper thus explores {\em multi-task oriented generative modeling} (MGM), by coupling a discriminative multi-task network with a generative network. To make them cooperate with each other, a straightforward solution would be to synthesize both RGB images and corresponding {\em pixel-level annotations} (\eg, pixel-wise class labels for semantic segmentation and depth map for depth estimation). In the single task scenario, existing work trains a separate generative model to synthesize paired pixel-level labeled data~\cite{sandfort2019data, choi2019self} and produce an augmented set. However, the quality and distribution of the generated annotations are not guaranteed. Moreover, these models are still highly task-dependant, and extending them to multi-task scenarios becomes difficult. A natural question then is: Do we actually need to synthesize paired image and multi-annotation data to be useful for multi-task visual learning?

Our MGM addresses this question by proposing a {\em general} framework that uses synthesized images paired with {\em only weak annotations} (\ie, image-level scene labels) to facilitate multiple visual tasks. Our key insight is to introduce {\em auxiliary discriminative tasks} that (i) only require image-level annotation or no annotation, and (ii) correlate with the original multiple tasks of interest. To this end, as additional components of the discriminative multi-task network, we introduce a {\em refinement} network and a {\em self-supervision} network that satisfies these properties. Through joint training, the discriminative network {\em explicitly} guides the image synthesis process. The generative network also contributes to further refining the shared feature representation. Meanwhile, the synthesized images of the generative network are used as additional training data for the discriminative network.%

In more detail, the refinement network performs scene classification on the basis of the multi-task network predictions, which requires only scene labels for images. The self-supervision network can be operationalized on both real and synthesized images without reliance on annotations. With these two modules, our MGM is able to learn from both (pixel-wise) fully-annotated real images and synthesized (image-level) weakly labeled images. We instantiate MGM with the state-of-the-art encoder-decoder based multi-task network~\cite{zamir2018taskonomy}, self-attention GAN~\cite{zhang2019self}, and contrastive learning-based self-supervision network~\cite{chen2020simple}. Note that our framework is {\em agnostic to the choice of these model components}.

We evaluate our approach on standard multi-task benchmarks, including the NYUv2~\cite{Silberman:ECCV12} and Taskonomy~\cite{zamir2018taskonomy} datasets. Consistent with the previous work~\cite{sun2019adashare,standley2019tasks}, we focus on three tasks of great practical importance: semantic segmentation, depth estimation, and normal prediction. The evaluation shows that our MGM consistently outperforms state-of-the-art multi-task approaches by large margins, almost reaching the {\em performance upper-bound} that trains with weakly annotated {\em real} images. Finally, we show the scalability of our approach to more visual tasks.

{\bf Our contributions} are three-folds. (1) We introduce a novel problem of multi-task oriented generative modeling, with the aim of leveraging generative networks to facilitate multi-task visual learning. (2) We propose MGM --- a general framework that effectively uses synthesized images paired with only weak annotations for multi-tasks, through a refinement network and a self-supervision network. (3) We show that jointly training of generative and discriminative multi-task networks under our MGM framework significantly improves the performance of all the tasks.

\begin{figure*}[t]
    \centering
    \includegraphics[width = 0.95 \linewidth]{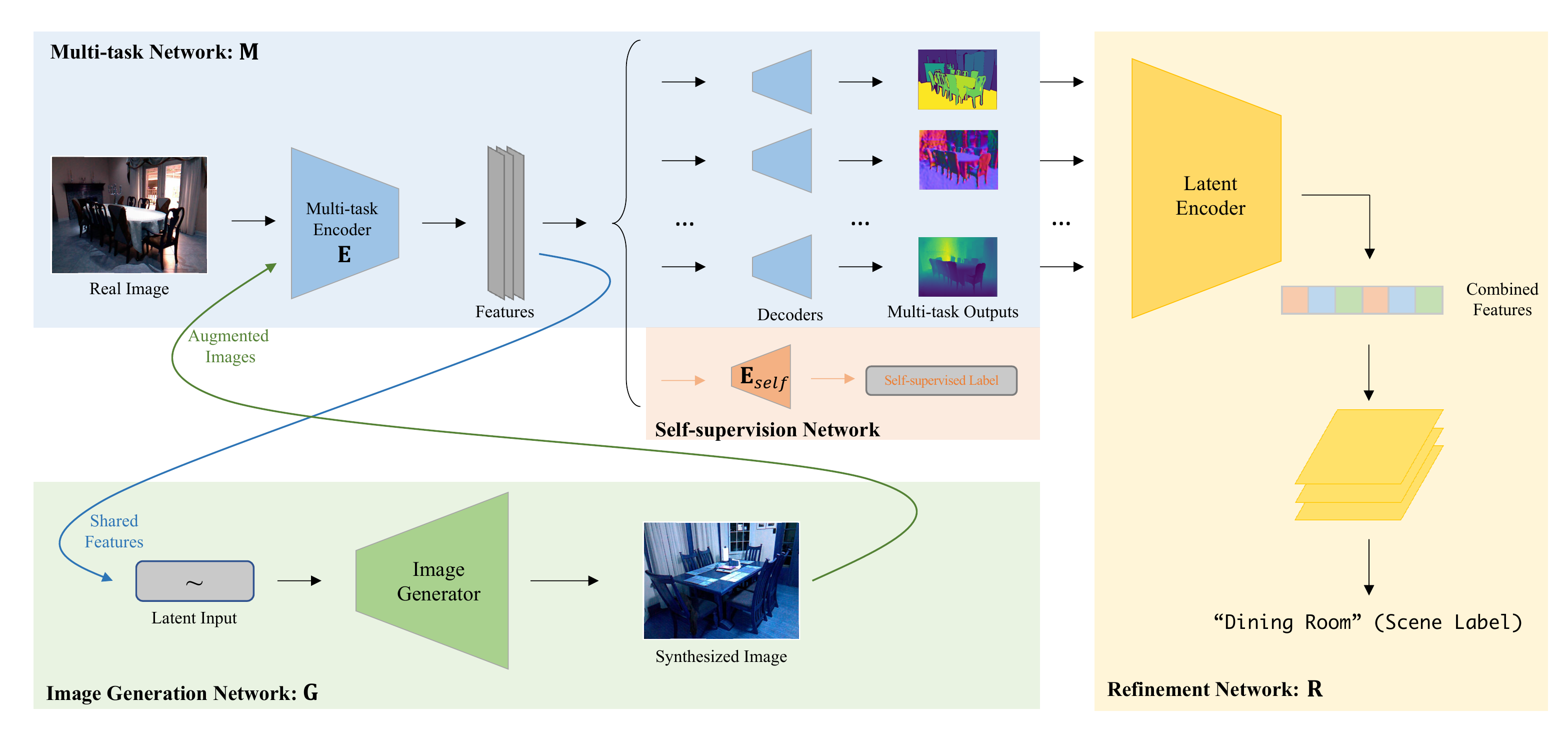}
    \vspace{-18 pt}
    \caption{Architecture of our proposed multi-task oriented generative modeling (MGM) framework. There are four main components in the framework: Multi-task network to address the target multiple pixel-level prediction tasks; self-supervision network to facilitate representation learning using images without any annotation; refinement network to perform scene classification using weak annotation; image generation network to synthesize useful images that benefit multiple tasks.}
    \vspace{-15 pt}
    \label{fig:model}
\end{figure*}

\section{Pilot Study}
\label{sec:pilot}
This pilot study provides some initial experimentation, which shows the importance of our proposed problem (multi-task oriented generative modeling) and motivates the following development of our method. Specifically, we show that {\em directly using images synthesized by an off-the-shelf generative model that is trained with the realistic objective are not helpful for downstream pixel-level perception tasks}.

\textbf{Experimental Design:} For ease of analysis, here we focus on a single task -- semantic segmentation, and use the Tiny-Taskonomy dataset~\cite{zamir2018taskonomy}. The dataset split and evaluation metric are the same as our main experiments (See Section~\ref{sec:exp} for details). We train a self-attention GAN~\cite{zhang2019self} on Tiny-Taskonomy, and use it to generate the same number of synthesized images as the real images to augment the training set.

\textbf{How to generate pixel-level annotations?} One remaining question is how to generate pixel-level annotations for these synthesized images produced by the self-attention GAN. While prior work has explored synthesizing both images and their pixel-level annotations for some specific tasks~\cite{sandfort2019data, choi2019self}, these annotations are still not reliable. For ease of analysis, in this study, we factor out the effect of annotations and assume that we have an {\em oracle annotator}. We use the annotator from Taskonomy~\cite{zamir2018taskonomy}, which is a state-of-the-art large fully supervised semantic segmentation network. In fact, the ground-truth of semantic segmentation on Taskonomy is produced as the output of this network rather than labeled by humans. By doing so, we ensure that the annotations of the synthesized images are ``accurate'' (in the sense that they are consistent with how the real images are labeled).

\textbf{Comparisons:} \textbf{Single-Task (ST)} model is our baseline which follows the architecture of Taskonomy single task network. ST is trained on real images only. \textbf{$\text{ST}_\mathrm{G}$} is the ST model trained on the augmented set. Table~\ref{tab:pilot} reports the results of these two models, and \textbf{$\text{ST}_\mathrm{G}$} is worse than \textbf{ST}.

\textbf{Does synthesized images help downstream tasks?} From Table~\ref{tab:pilot}, the answer is \textbf{NO}. Even though the images are synthesized by one of the state-of-the-art generative models and are labeled by the oracle annotator, they still cannot benefit the downstream task. This is probably because these images are synthesized off the shelf {\em without} ``knowing'' the downstream task. {\em Our key insight} then is that we need to explicitly use the downstream task objective to guide the image synthesis process. Moreover, here we focused on a single task and assumed that we had the oracle annotations. However, an oracle annotator is difficult to obtain in practice, especially for multiple tasks. Also, existing work cannot synthesize paired images and pixel-level annotations for multiple tasks~\cite{sandfort2019data, choi2019self}. To overcome these challenges, in what follows we demonstrate how to facilitate multiple visual tasks with synthesize images that only need image-level scene labels.
\begin{table}[t]
    \centering
    \begin{tabular}{c|cc}
    \hline
        Model & ST & $\text{ST}_\mathrm{G}$ \\ \hline
        mLoss ($\downarrow$) & 0.111 & 0.148\\ \hline
    \end{tabular}
    \vspace{-5pt}
    \caption{Pilot experiment for semantic segmentation on the Tiny-Taskonomy dataset. Directly using images synthesized by an off-the-shelf generative model (self-attention GAN) may hurt the performance on the downstream task. ST: single-task semantic segmentation model trained on real images only; $\text{ST}_\mathrm{G}$: the same model trained on both real and synthesized images.}
    \label{tab:pilot}
    \vspace{-23pt}
\end{table}

\vspace{-5pt}
\section{Method}
\label{sec:method}

We propose multi-task oriented generative modeling (MGM) to leverage generative networks for multi-task visual learning, as summarized in Figure~\ref{fig:model}. In this section, we first formalize the novel problem setting of MGM. Then, we explain the general framework and an instantiation of the MGM model with state-of-the-art multi-task learning and image generation approaches. Finally, we discuss the detailed training strategy for the framework. 

\vspace{-5pt}
\subsection{Problem Setting}
\label{sec:problem}
\textbf{Multi-task discriminative learning:} Given a set of $n$ visual tasks $\mathcal{T} = \{T_1,T_2,\cdots,T_n\}$, we aim to learn a discriminative multi-task model $\mathbf{M}$ that is able to address all of these tasks simultaneously: $\mathbf{M}(x)\rightarrow \widehat{\boldsymbol{y}}=(\widehat{y}^1,\widehat{y}^2,\cdots,\widehat{y}^n)$, where $x$ is an input image and $\widehat{y}^i$ is the prediction for task $T_i$. Here we focus on the type of per-pixel level prediction tasks (\eg, semantic segmentation or depth estimation). We treat image classification as a special task, which provides global semantic description (\ie, scene labels) of images and only requires image-level category annotation $c$. Therefore, the set of fully annotated real data is denoted as $\mathcal{S}_\mathrm{real} = \{(x_j, y_j^1, y_j^2,\cdots,y_j^n,c_j)\}$.

\textbf{Generative learning:}
Meanwhile, we aim to learn a generative model $\mathbf{G}$ that produces a set of synthesized data but with only corresponding image-level scene labels (weak annotation): $\mathbf{G}(c,z) \rightarrow \widetilde{x}$, where $z$ is a random input, and $\widetilde{x}$ is a synthesized image. The scene label of $\widetilde{x}$ is denoted as $\widetilde{c} = c$. We denote the set of synthesized images and their corresponding scene labels as $\widetilde{\mathcal{S}}_\mathrm{syn} = \{(\widetilde{x}_k,\widetilde{c}_k)\}$.

\textbf{Cooperation between discriminative and generative learning:} Our objective is that the discriminative model $\mathbf{M}$ and the generative model $\mathbf{G}$ cooperate with each other to improve the performance on the multiple visual tasks $\mathcal{T}$. During the whole process, the full model only gets access to the real fully-labeled data $\mathcal{S}_\mathrm{real}$, then the generative network is trained to produce the synthesized set $\widetilde{\mathcal{S}}_\mathrm{syn}$. Finally, $\mathbf{M}$ effectively learns from both $\mathcal{S}_\mathrm{real}$ and $\widetilde{\mathcal{S}}_\mathrm{syn}$.
Note that, unlike most of the existing work on image generation~\cite{brock2018large,zhang2019self}, {\em we do not focus on the visual realism of the synthesized images} $\widetilde{x}$. Instead, we hope $\mathbf{G}$ to capture the underlying mechanism that benefits $\mathbf{M}$.

\subsection{Framework and Architecture}
\label{sec:framework}
Figure~\ref{fig:model} shows the architecture of our proposed MGM framework. It contains four components: the main multi-task discriminative network $\mathbf{M}$, the image generation network $\mathbf{G}$, the refinement network $\mathbf{R}$, and the self-supervision network. By introducing the refinement network and the self-supervision network, the full model can leverage both fully-labeled real images and weakly-labeled synthesized images to facilitate the learning of latent feature representation. These two networks thus allow $\mathbf{M}$ and $\mathbf{R}$ to better cooperate with each other.
Notice that our MGM is a {\em model-agnostic} framework, and here we instantiate its components with state-of-the-art models. In the ablation study (Sec.~\ref{sec:ablation}) and the supplementary material, we show that our MGM works well with different choices of the model components.

\textbf{Multi-task Network (\textbf{M}):} The multi-task network aims to make predictions for multiple target tasks based on an input image. Consistent with the most recent work on multi-task learning, we instantiate an encoder-decoder based architecture~\cite{zamir2018taskonomy,zhang2019self,sun2019adashare}. Considering the trade-off between model complexity and performance, we use a shared encoder $\mathbf{E}$ to extract features from input images, and individual decoders for each target task. We 
adopt a ResNet-18~\cite{he2016deep} for the encoder and symmetric transposed decoders following~\cite{zamir2018taskonomy}. For each task, we have its own loss function to update the corresponding decoder and the shared encoder. 

\textbf{Image Generation Network (\textbf{G}):} The generative model $\mathbf{G}$ is a variant of generative adversarial networks (GANs). We include the generator in our framework, but this module also has a discriminator during its own training. $\mathbf{G}$ takes as input a latent vector $z$ and a category label $c$, and synthesizes an image belonging to category $c$. Considering the trade-off between performance and training cost, we instantiate $\mathbf{G}$ with self-attention generative adversarial network (SAGAN)~\cite{zhang2019self}. We achieve conditional image generation by applying conditional batch normalization (CBN) layers~\cite{de2017modulating}:
\vspace{-3pt}
\begin{equation}
    \mathrm{CBN}\left(f_{i, c, h, w} \mid \gamma_{c}, \beta_{c}\right)=\gamma_{c} \frac{f_{i, c, w, h}-\mathrm{E}\left[f_{\cdot, c, \cdot, \cdot}\right]}{\sqrt{\operatorname{Var}\left[f_{\cdot, c, \cdot, \cdot}\right]+\epsilon}}+\beta_{c},
\end{equation}
where $f_{i,c,h,w}$ is an extracted $c$-channel 2D feature for the $i$-th sample, and $\epsilon$ is a small value to avoid collapse. $\gamma_c$ and $\beta_c$ are two parameters to control the mean and variance of the normalization, which are learned by the model for each class. We use hinge loss for the adversarial training. Notice that the proposed framework is flexible with different generative models, and we also show the effectiveness of using DCGAN~\cite{radford2015unsupervised} in the supplementary.

\textbf{Refinement Network (\textbf{R}):} As one of our key contributions, we introduce the refinement network $\mathbf{R}$ to further refine the shared representation using the global scene category labels. \textbf{R} takes the predictions of the multi-task network as input and predicts the category label of the input image. Importantly, because $\mathbf{R}$ only requires category labels, it can be effortlessly operationalized on the ``weakly-annotated'' synthesized images. Meanwhile, $\mathbf{R}$ also enforces the semantic consistency of the synthesized images with \textbf{G}.

We apply an algorithm inspired by Expectation-Maximum (EM)~\cite{dempster1977maximum} to train the refinement network \textbf{R}. For the fully-annotated real images $(x,\boldsymbol{y},c)$, we use the scene classification loss to update $\mathbf{R}$ and refine the encoder $\mathbf{E}$ in the multi-task network $\mathbf{M}$. Then for the synthesized images $(\widetilde{x},\widetilde{c})$, since their multi-task predictions produced by $\mathbf{M}$ might not be reliable, we only refine $\mathbf{E}$ with $\mathbf{R}$ frozen using the scene classification loss. Through refining the share feature representation with the synthesized images, this process also provides implicit guidance to the image generation network.

More specifically, we model the whole multi-task network and refinement network as a joint probability graph:
\vspace{-3pt}
\begin{equation}
P(x, \boldsymbol{y}, c; \theta,\theta^\prime)=P(x)\left(\prod_{i=1}^{n} P\left(y^{i} \mid x; \theta\right)\right) P(c \mid \boldsymbol{y}; \theta^\prime),
\end{equation}
where $x$ is an input image, $\boldsymbol{y}$ is the vector of multi-task predictions, $c$ is the scene label, $\theta$ is the vector of parameters of the multi-task network, and $\theta^\prime$ is the vector of parameters of the refinement network. The parameters $\theta$ and $\theta^\prime$ are learned to maximize the joint probability. For data samples in $\mathcal{S}_\mathrm{real}$, we maximize the joint probability and update $\theta^\prime$ to train the refinement network.
\vspace{-3pt}
\begin{equation}
    \theta^{\prime \star} = \underset{\theta^\prime}{\operatorname{argmax}}~P(\widetilde{c}_k \mid \boldsymbol{y};\theta^\prime).
    \vspace{-12 pt}
\end{equation}
For data samples in $\widetilde{\mathcal{S}}_\mathrm{syn}$, we update the parameters of $\mathbf{M}$ ($\theta$) in an EM-like manner. During the $\textbf{E}$ step, we estimate the latent multi-task ground-truth by:
\vspace{-3pt}
\begin{equation}
    {\boldsymbol{y}^\dagger}=\underset{\boldsymbol{y}}{\operatorname{argmax}}~P\left(\boldsymbol{y} \mid \widetilde{x}_k ; \theta \right) P(\widetilde{c}_k \mid \boldsymbol{y};\theta^\prime).
\label{eq:mstep}
\vspace{-1 pt}
\end{equation}
Then for the $\textbf{M}$ step, we back-propagate the error between ${\boldsymbol{y}}^\dagger$ and $\widehat{\boldsymbol{y}}$ (the multi-task predictions) to the multi-task encoder. 
\begin{equation}
\vspace{-3 pt}
    \theta^{\star} = \underset{\theta}{\operatorname{argmax}}~P\left(\boldsymbol{y}^\dagger \mid \widetilde{x}_k ; \theta \right).
    \vspace{-4 pt}
\end{equation}
We use cross-entropy as the classification loss function.

\textbf{Self-supervision Network:} The self-supervision network facilitates the representation learning of the encoder $\mathbf{E}$ by performing self-supervised learning tasks on images without any annotation so that can be operationalized on both real and synthesized images. We modify SimCLR~\cite{chen2020simple}, one of the state-of-the-art approaches, as our self-supervision network. 

This network contains an additional embedding network $\mathbf{E}_\mathrm{self}$, working on the output of the multi-task encoder $\mathbf{E}$, to obtain a 1D latent feature of the input image: $\mu = \mathbf{E}_\mathrm{self}(\mathbf{E}(x))$. Then, it performs contrastive learning with these latent vectors. Specifically, given a minibatch of $N$ images, this network first randomly samples two transformed views of each source image as augmented images (See supplementary material for the detailed transformations), resulting in $2N$ augmented images. For each augmented image, there is only one pair of positive augmented examples from the same source image, and other $2(N-1)$ negative pairs. Then the network jointly minimizes the distance of positive pairs and maximizes the distance of negative pairs in the latent space, through the normalized temperature-scaled cross-entropy (\textit{NT-Xent})) loss~\cite{chen2020simple}:    
\begin{equation}
    \ell_{i, j}=-\log \frac{\exp \left(\operatorname{dis}\left({\mu}_{i}, {\mu}_{j}\right) / \tau\right)}{\sum_{k=1}^{2 N} \mathbbm{1}_{[k \neq i]} \exp \left(\operatorname{dis}\left({\mu}_{i}, {\mu}_{k}\right) / \tau\right)},
\label{eq:ntxent}
\vspace{-7 pt}
\end{equation}
where $\ell_{i, j}$ is the \textit{NT-Xent} loss for a positive pair of examples in the latent space $(\mu_i, \mu_j)$. $\mathbbm{1}_{[k \neq i]} \in {0, 1}$ is an indicator function evaluating to 1 if $k\neq i$, and $\tau$ is a temperature parameter. $\operatorname{dis}\left({\mu}_{i}, \mu_{j}\right)$ is a distance function, and we use cosine distance following~\cite{chen2020simple}. This loss is further back-propagated to refine the multi-task encoder $\mathbf{E}$. Notice that other types of self-supervised tasks are applicable as well. To demonstrate this, in Sec.~\ref{sec:ablation} we also report the result with another task --- image reconstruction.

\subsection{Interaction Among Networks}
\label{sec:inter}
\begin{figure}[t]
    \centering
    \includegraphics[width = 0.9 \linewidth]{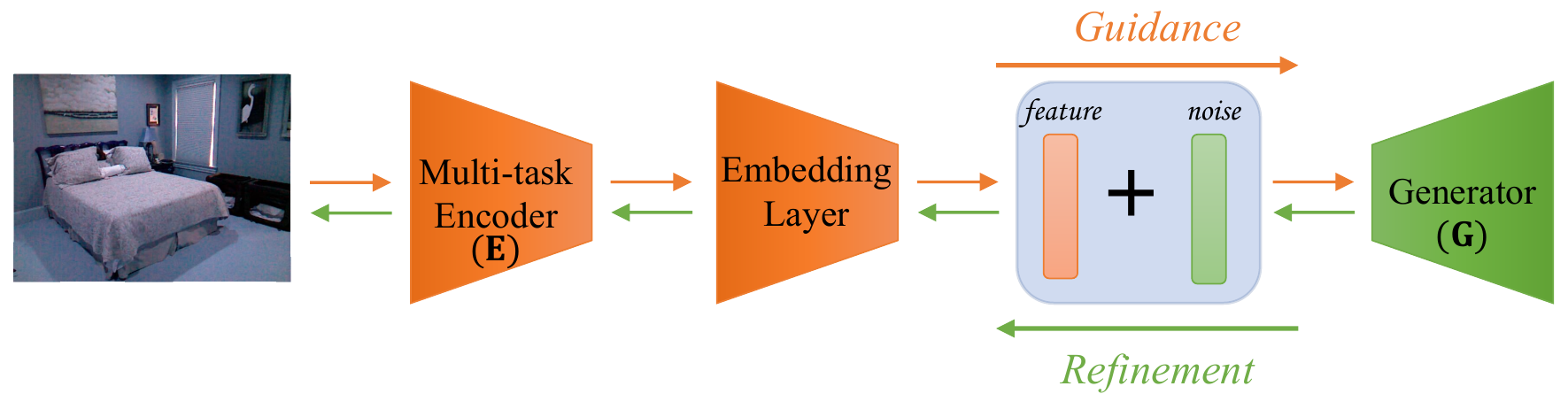}
    \vspace{-12 pt}
    \caption{Joint training of the multi-task network and the image generation network. The multi-task network provides useful feature representation to guide the image generation process, while the generation network refines the shared representation through back-propagation.}
    \vspace{-20 pt}
    \label{fig:joint}
\end{figure}
\textbf{Cooperation Through Joint Training:} We propose a simple but effective joint training algorithm shown in Figure~\ref{fig:joint}. The image generation network $\mathbf{G}$ takes the transferred feature representation of the multi-task encoder $\mathbf{E}$, added with some Gaussian noise, as the latent input $z$ to conduct conditional image generation. Hence, the generation network obtains {\em additional, explicit guidance} (\ie, extra effective features) from the multi-task network to facilitate the generation of ``better images''--- images that may not look more realistic but are more useful for the multiple target tasks. Then, the generation error of $\mathbf{G}$ will be back-propagated to $\mathbf{E}$ to further refine the shared representation. This process can be also viewed as introducing image generation as an additional task in the multi-task learning framework. 

\textbf{Training Procedure:} We describe the procedure in Algorithm~\ref{alg1} and further explain it in the supplementary material.

\begin{algorithm}[t]
\SetAlgoLined
\begin{small}
\caption{The training procedure of MGM}
\label{alg1}
\textbf{Initialization:} \\
 $max\_epoch$: Maximum epoch for the training;\\
 $M$: Multi-task Network, $G$: Image Generation Network;\\
 $E$: Multi-task Encoder, $R$: the Refinement Network; \\
 $E_\mathrm{self}$: Self-supervision Network Encoder;\\
 $N$: minibatch size;\\
 \For{epoch $\leftarrow 1 \ \KwTo \ max\_epoch$}{
 Split $\mathcal{S}_\mathrm{real}$ into minibatches with size $N$: $\mathcal{S}_\mathrm{mini}$; \\
 \For{$(x,\boldsymbol{y},c) \in \mathcal{S}_\mathrm{mini}$}{
 $\hat{\boldsymbol{y}} = \mathbf{M}(x)$ \\
 $\mathcal{L}_{multi}(\boldsymbol{y},\hat{\boldsymbol{y}}) \rightarrow$ update $\mathbf{M}$; \\
 $\hat{c} = \mathbf{R}(\hat{\boldsymbol{y}})$ \\
 $\mathcal{L}_{CE}(c,\hat{c})\rightarrow$ update $\mathbf{R}$, $\mathbf{E}$; \\
 Sample $2N$ augmented images $x_\mathrm{aug}$ \\
 $\mathcal{L}_{NT-Xent}(x_\mathrm{aug}) \rightarrow$ update $\mathbf{E}$, $\mathbf{E}_\mathrm{self}$; \\
 Use $\mathcal{L}_{GAN}$ to train $\mathbf{G}$; \\
 $(\widetilde{x},\widetilde{c}) = \mathbf{G}(\mathbf{x},c)$\\
 $\mathcal{L}_{CE}(\widetilde{c},\mathbf{R}(\mathbf{M}(\widetilde{x}))) \rightarrow$ update $\mathbf{E}$ \\
 Sample $2N$ augmented images for the synthesized data $\widetilde{x}_\mathrm{aug}$ \\
 $\mathcal{L}_{NT-Xent}(\widetilde{x}_\mathrm{aug}) \rightarrow$ update $\mathbf{E}$.
 }
 }
 \end{small}
\end{algorithm}

\section{Experiments}
\label{sec:exp}

\begin{table*}[t]
    \centering
    \small
    \resizebox{\linewidth}{!}{
    \begin{tabular}{c|c|ccc|ccc|c|ccc|c}
    \hline
      & Data Setting & \multicolumn{3}{c|}{100\% Data Setting} &   \multicolumn{4}{c|}{50\% Data Setting} & \multicolumn{4}{c}{25\% Data Setting}\\\cline{2-13}
      & Models & ST & MT & MGM  & ST & MT & MGM &\it $\text{MGM}_\mathrm{r}$ & ST & MT & MGM &\it $\text{MGM}_\mathrm{r}$ \\ \hline 
      \multirow{3}{*}{\shortstack{NYU \\ v2}} 
    & SS-mIOU($\uparrow$) & \tabincell{c}{0.249 \\ $\pm$ 0.008} &  \tabincell{c}{0.256 \\ $\pm$ 0.005} & \tabincell{c}{\bf 0.264 \\ $\bf \pm$ \bf 0.005} & \tabincell{c}{0.230 \\ $\pm$ 0.009} &    \tabincell{c}{0.237 \\ $\pm$ 0.006}  & \tabincell{c}{\bf 0.251 \\ $\bf \pm$ \bf 0.005} & \it \tabincell{c}{\it 0.258 \\ $\it \pm$ \it 0.004} & \tabincell{c}{0.199 \\ $\pm$ 0.004} & \tabincell{c}{0.207 \\ $\pm$ 0.007}  & \tabincell{c}{\bf 0.229 \\ $\bf \pm$ \bf 0.004} & \it \tabincell{c}{\it 0.231 \\ $\it \pm$ \it 0.005} \\ \cline{2-13}
    
    & DE-mABSE($\downarrow$)& \tabincell{c}{0.748 \\ $\pm$ 0.019} &  \tabincell{c}{0.708 \\ $\pm$ 0.021} & \tabincell{c}{\bf 0.698 \\ $\bf \pm$ \bf 0.014} & \tabincell{c}{0.837 \\ $\pm$ 0.017} &    \tabincell{c}{0.819 \\ $\pm$ 0.018}  & \tabincell{c}{\bf 0.734 \\ $\bf \pm$ \bf 0.011} & \it \tabincell{c}{\it 0.723 \\ $\it \pm$ \it 0.010} & \tabincell{c}{0.908 \\ $\pm$ 0.017} & \tabincell{c}{0.874 \\ $\pm$ 0.015}  & \tabincell{c}{\bf 0.844 \\ $\bf \pm$ \bf 0.011} & \it \tabincell{c}{\it 0.821 \\ $\it \pm$ \it 0.009} \\ \cline{2-13}
    
    & SN-mAD($\downarrow$) & \tabincell{c}{0.273 \\ $\pm$ 0.06} &  \tabincell{c}{0.283 \\ $\pm$ 0.008} & \tabincell{c}{\bf 0.255 \\ $\bf \pm$ \bf 0.010} & \tabincell{c}{0.309 \\ $\pm$ 0.008} &    \tabincell{c}{0.291 \\ $\pm$ 0.010}  & \tabincell{c}{\bf 0.273 \\ $\bf \pm$ \bf 0.009} & \it \tabincell{c}{\it 0.270 \\ $\it \pm$ \it 0.006} & \tabincell{c}{0.312 \\ $\pm$ 0.007} & \tabincell{c}{0.296 \\ $\pm$ 0.007}  & \tabincell{c}{\bf 0.277 \\ $\bf \pm$ \bf 0.006} & \it \tabincell{c}{\it 0.274 \\ $\it \pm$ \it 0.005} \\ \hline \hline 
    
      \multirow{3}{*}{\shortstack{\footnotesize Tiny \\ Taskonomy}} 
      & SS-mLoss($\downarrow$) & \tabincell{c}{0.111 \\ $\pm$ 0.002} &  \tabincell{c}{0.137 \\ $\pm$ 0.003} & \tabincell{c}{\bf 0.106 \\ $\bf \pm$ \bf 0.003} & \tabincell{c}{0.120 \\ $\pm$ 0.003} &    \tabincell{c}{0.138 \\ $\pm$ 0.002}  & \tabincell{c}{\bf 0.114 \\ $\bf \pm$ \bf 0.003} & \it \tabincell{c}{\it 0.112 \\ $\it \pm$ \it 0.002} & \tabincell{c}{0.119 \\ $\pm$ 0.003} & \tabincell{c}{0.141 \\ $\pm$ 0.002}  & \tabincell{c}{\bf 0.117 \\ $\bf \pm$ \bf 0.002} & \it \tabincell{c}{\it 0.115 \\ $\it \pm$ \it 0.002} \\ \cline{2-13}
      
      & DE-mLoss($\downarrow$) & \tabincell{c}{1.716 \\ $\pm$ 0.006} &  \tabincell{c}{1.584 \\ $\pm$ 0.008} & \tabincell{c}{\bf 1.472 \\ $\bf \pm$ \bf 0.006} & \tabincell{c}{1.768 \\ $\pm$ 0.007} &    \tabincell{c}{1.595 \\ $\pm$ 0.009}  & \tabincell{c}{\bf 1.499 \\ $\bf \pm$ \bf 0.008} & \it \tabincell{c}{\it 1.378 \\ $\it \pm$ \it 0.007} & \tabincell{c}{1.795 \\ $\pm$ 0.010} & \tabincell{c}{1.692 \\ $\pm$ 0.008}  & \tabincell{c}{\bf 1.585 \\ $\bf \pm$ \bf 0.009} & \it \tabincell{c}{\it 1.580 \\ $\it \pm$ \it 0.008} \\ \cline{2-13}
      
      & SN-mLoss($\downarrow$) & \tabincell{c}{0.155 \\ $\pm$ 0.003} &  \tabincell{c}{0.153 \\ $\pm$ 0.003} & \tabincell{c}{\bf 0.145 \\ $\bf \pm$ \bf 0.002} & \tabincell{c}{0.157 \\ $\pm$ 0.002} &    \tabincell{c}{0.156 \\ $\pm$ 0.002}  & \tabincell{c}{\bf 0.147 \\ $\bf \pm$ \bf 0.002} & \it \tabincell{c}{\it 0.140 \\ $\it \pm$ \it 0.001} & \tabincell{c}{0.154 \\ $\pm$ 0.002} & \tabincell{c}{0.152 \\ $\pm$ 0.002}  & \tabincell{c}{\bf 0.148 \\ $\bf \pm$ \bf 0.003} & \it \tabincell{c}{\it 0.142 \\ $\it \pm$ \it 0.002} \\ \hline
    \end{tabular}}
    \vspace{-10 pt}
    \caption{Main results (mean $\pm$ std) on the NYUv2 and Tiny-Taskonomy datasets. SS: semantic  segmentation; DE: depth estimation; SN: surface  normal  prediction. $\uparrow$ means higher is better; $\downarrow$ means lower is better. We use different metrics on the two datasets, following existing protocol. Our MGM consistently and significantly outperforms both single-task (ST) and multi-task (MT) baselines, {\em even reaching the performance upper-bound of training with weakly annotated real images} ($\text{MGM}_\mathrm{r}$).}
    \label{tab:main}
    \vspace{-9 pt}
\end{table*}

\begin{table*}[t]
    \centering
    \small
    \begin{tabular}{c|c|ccc|ccc|ccc}
    \hline
       & Data Setting & \multicolumn{3}{c|}{100\% Data Setting} &   \multicolumn{3}{c|}{50\% Data Setting} & \multicolumn{3}{c}{25\% Data Setting}\\\cline{2-11}
      & Models & $\text{MGM}_\mathrm{/G}$ & $\text{MGM}_\mathrm{/j}$ & MGM  & $\text{MGM}_\mathrm{/G}$ & $\text{MGM}_\mathrm{/j}$ & MGM & $\text{MGM}_\mathrm{/G}$ & $\text{MGM}_\mathrm{/j}$ & MGM  \\ \hline 
      \multirow{3}{*}{\shortstack{NYU \\ v2}} & 
      SS-mIOU($\uparrow$) & 0.261  & 0.262  & 0.264  & 0.243 & 0.243 & 0.251  & 0.215 & 0.220 & 0.229  \\
      & DE-mABSE($\downarrow$)& 0.707 & 0.701 & 0.698  & 0.799 & 0.763 & 0.734  & 0.868 & 0.860 & 0.844   \\
      & SN-mAD($\downarrow$) & 0.262 & 0.259 & 0.255 & 0.287  & 0.281 & 0.273 & 0.292 & 0.286 & 0.277  \\ \hline \hline
      \multirow{3}{*}{\shortstack{Tiny \\ Taskonomy}} 
      & SS-mLoss($\downarrow$) & 0.108 & 0.108 & 0.106 & 0.116 & 0.115 & 0.114 & 0.119 & 0.121 & 0.117  \\
      & DE-mLoss($\downarrow$) & 1.491 & 1.488 & 1.472 & 1.527 & 1.523 & 1.499 & 1.636 & 1.616 & 1.585 \\
      & SN-mLoss($\downarrow$) & 0.151 & 0.151 & 0.145 & 0.153 & 0.152 & 0.147  & 0.154 & 0.152 & 0.148  \\\hline 
      
    \end{tabular}
    \vspace{-8 pt}
    \caption{Comparison of our MGM model with its variants. $\text{MGM}_\mathrm{/G}$: {\em without} generating synthesized images; $\text{MGM}_\mathrm{/j}$: {\em without} joint learning. Our MGM outperforms single-task and multi-task baselines {\em even without synthesized data}, showing its effectiveness as a general multi-task learning framework. The model performance further improves with joint learning.}
    \label{tab:components}
    \vspace{-16 pt}
\end{table*}

To evaluate our proposed MGM model and investigate the impact of each component, we conduct a variety of experiments on two standard multi-task learning datasets. We also perform detailed analysis and ablation studies here. 
\subsection{Datasets and Compared Methods}
\textbf{Datasets:} Following the work of \cite{sun2019adashare} and \cite{standley2019tasks}, we mainly focus on three representative visual tasks in the main experiments: semantic segmentation (SS), surface normal prediction (SN), and depth estimation (DE). At the end of this section, we will show that our approach is scalable to an additional number of tasks. We evaluate all the models on two widely-benchmarked datasets: \textbf{NYUv2}~\cite{Silberman:ECCV12,eigen2015predicting} containing 1,449 images with 40 types of objects~\cite{gupta2013perceptual}; \textbf{Tiny-Taskonomy} which is the standard tiny split of the Taskonomy dataset~\cite{zamir2018taskonomy}.

Since a certain amount of images for each category is required to train a generative network, we keep the images of the top 35 scene categories on Tiny-Taskonomy, with each one consisting of more than 1,000 images. This resulting dataset contains 358,426 images in total. For NYUv2, we randomly select 1,049 images as the full training set and 200 images each as the validation/test set. For Tiny-Taskonomy, we randomly pick 80\% of the whole set as the full training set and 10\% each as the validation/test set. 

\textbf{Compared Methods:} We mainly focus on our comparison with two state-of-the-art discriminative baselines: \textbf{Single-Task (ST)} model follows the architecture of Taskonomy single task network~\cite{zamir2018taskonomy}, and address each task individually; \textbf{Multi-Task (MT)} model refers to the sub-network for the three tasks of interest in~\cite{standley2019tasks}. These two baselines can be viewed as using our multi-task network without the proposed refinement, self-supervision, and generation networks. Note that {\em our work is the first that introduces generative modeling for multi-task learning, and there is no existing baseline in this direction}.

Our \textbf{MGM} is the full model trained with both fully-labeled {\em real} data and weakly-labeled {\em synthesized} data, which are produced by the generation network through joint training. In addition, to further validate the effectiveness of our \textbf{MGM} model, we consider its variant model \textbf{$\text{MGM}_\mathrm{r}$} that is trained with both fully and weakly labeled {\em real} data. \textbf{$\text{MGM}_\mathrm{r}$} is used to show {\em the performance upper bound} in the semi-supervised learning scenario, where the synthesized images are replaced by the real images in the dataset. The resolution is set to 128 for all the experiments. For all the compared methods, we use a ResNet-18 like architecture to build the encoder and use the standard decoder architecture of Taskonomy~\cite{zamir2018taskonomy}.

\textbf{Data Settings:}~
We conduct experiments with three different data settings: (1) 100\% data setting; (2) 50\% data setting; and (3) 25\% data setting. For each setting, we use 100\%, 50\%, or 25\% of the entire labeled training set to train the model. For \textbf{$\text{MGM}_\mathrm{r}$}, we add another 50\% or 25\% of weakly-labeled real data in the last two settings. For \textbf{MGM}, we include the same number of weakly-labeled synthesized data in all three settings. 

\textbf{Evaluation Metrics:}~
For NYUv2, following the metrics in~\cite{eigen2015predicting,sun2019adashare}, we measure the mean Intersection-Over-Union (mIOU) for the semantic segmentation task, the mean Absolute Error (mABSE) for the depth estimation task, and the mean Angular Distance (mAD) for the surface normal estimation task. For Tiny-Taskonomy, we follow the evaluation metrics of previous work~\cite{zamir2018taskonomy,standley2019tasks,sun2019adashare} and report the averaged loss values on the test set.

\textbf{Implementation Details:}~
See the supplementary for the training details and the sensitivity of the hyper-parameters.

\subsection{Main Results}
\vspace{-2pt}
\textbf{Quantitative Results:} We run all the models for 5 times and report the averaged results and the standard deviation on the two datasets in Table~\ref{tab:main}. From this table, we have the following key observations that support the effectiveness of our approach which combines generative learning with discriminative learning. (1) Existing discriminative multi-task learning approaches may not consistently benefit all the three individual tasks. However, our MGM consistently and significantly outperforms both the single-task and multi-task baselines across all the scenarios. (2) By using weakly-labeled synthesized data, the results of our model in the 50\% data setting are even better than those of baselines in the 100\% data setting. (3) More interesting, the performance of our MGM is close to $\text{MGM}_\mathrm{r}$, which indicates that our synthesized images are {\em comparably useful} as real images for improving multiple visual perception tasks. The performance gap is especially minimal in the 25\% labeled data setting, suggesting that our proposed MGM model is, in particular, helpful for low-data regime.

\begin{figure*}[t]
    \centering
    \includegraphics[width = 0.85 \linewidth]{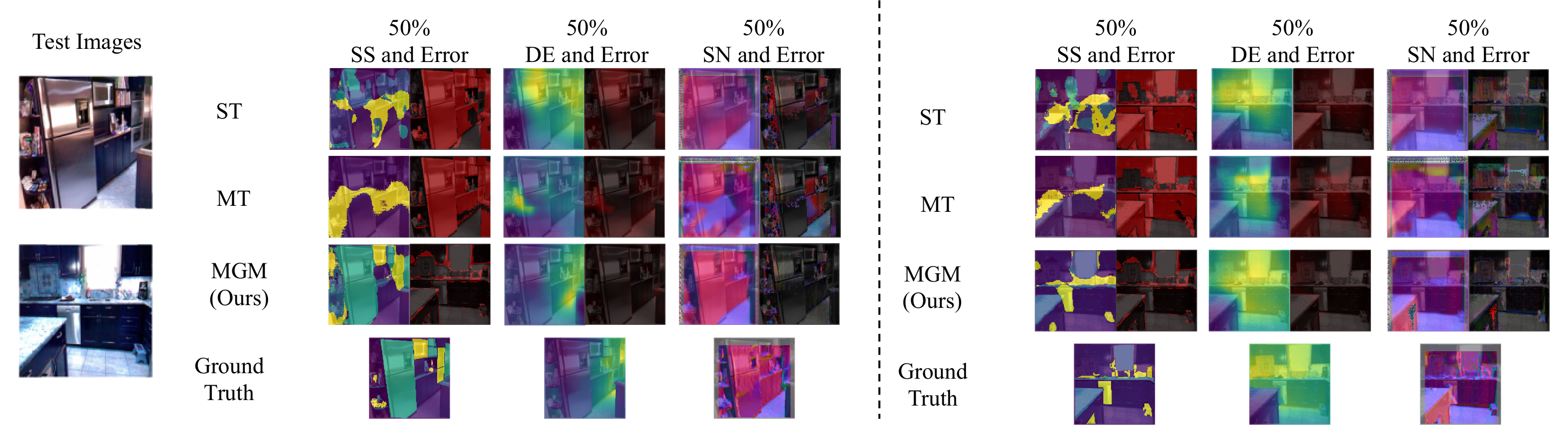}
    \vspace{-11 pt}
    \caption{Visualization and error comparison of the multi-task prediction outputs in the 50\% data setting. The prediction results of MGM is quite close to the ground-truth, significantly outperforming the state-of-the-art results.}
    \label{fig:quali}
    \vspace{-13 pt}
\end{figure*}

\vspace{-2pt}
\textbf{Qualitative Results:} We also visualize the prediction results on the three tasks for ST, MT, and MGM in the 50\% data setting in Figure~\ref{fig:quali}. While obvious defects can be found for all the baselines, the results of our proposed method are quite close to the ground-truth.

\vspace{-2pt}
\textbf{How Does Generative Modeling Benefit Multi-tasks?} To have a better understanding of how the generative modeling and joint learning mechanism benefit multi-task visual learning, we also consider two variants of our MGM model and evaluate their performance. \textbf{$\text{MGM}_\mathrm{/G}$} is the MGM model trained with $\mathcal{S}_\mathrm{real}$ only (without generative modeling), which shows the performance of our proposed multi-task learning framework in general (with the help from the auxiliary refinement and self-supervision networks), and helps to understand the gain of leveraging generative modeling. \textbf{$\text{MGM}_\mathrm{/j}$} is trained with synthesized images produced by a pre-trained SAGAN \textit{without} the joint training mechanism. Table~\ref{tab:components} shows the results on the two datasets. 

Combining the results of Tables~\ref{tab:components} and~\ref{tab:main}, we find: (1) MGM outperforms both ST and MT baseline even without generative modeling, indicating the benefit of the self-supervised task and the refinement network. (2) By introducing synthesized images that are trained separately, the multi-task performance slightly improves, which shows the effectiveness of involving generative modeling into multi-task discriminative learning, under the 
assistance of our refinement and self-supervision networks. (3) The joint learning mechanism further improves the cooperation between generative modeling and discriminative learning, thus enabling the generative model to better facilitate multi-task visual learning.

\vspace{-5pt}
\subsection{Ablation Study}
\label{sec:ablation}
For all the experiments in this section, models are trained in the 50\% data setting, unless specifically mentioned. 

\textbf{Impact of Parameters:} Introducing the refinement, self-supervision, and image generation networks also leads to more parameters. To validate that the performance improvements come from the novel design of our architecture rather than merely increasing the number of parameters, we provide two model variants as additional baselines: \textbf{$\textbf{ST}_\mathrm{l}$} and \textbf{$\textbf{MT}_\mathrm{l}$} use ResNet-34 as the encoder network and the corresponding decoder networks. These two networks have a 
similar amount of parameters as MGM. The top 4 rows in Table~\ref{tab:self} show that simply increasing the number of parameters cannot significantly boost performance.

\textbf{Impact of Self-supervision Task and Refinement Network:} Two important components of the proposed framework are the self-supervision task and the refinement network. We evaluate their impact individually in Table~\ref{tab:self}. $\text{MGM}_\mathrm{/self}$ is the model trained {\em without} the self-supervision task; $\text{MGM}_\mathrm{/refine}$ is the model {\em without} the refinement network; for $\text{MGM}_\mathrm{recon}$, we replace the SimCLR based self-supervision method with a weaker reconstruction task, and use Mean Square Error as the loss function.

We could see that the refinement network works better for the surface normal task, and the self-supervision task works better for the semantic segmentation task; they are complementary to each other, and combining them generally achieves the best performance. In addition, the model could still gain some benefit even when we use some weak self-supervision tasks like reconstruction, which indicates the generability and robustness of our MGM model.

\begin{table}[t]
    \centering
    \small
    \resizebox{\linewidth}{!}{
    \begin{tabular}{c|ccc}
    \hline
    Model & SS-mIOU ($\uparrow$) & DE-mABSE ($\downarrow$) & SN-mAD ($\downarrow$) \\
      \hline
      ST & 0.230 & 0.837 & 0.309\\
      MT & 0.237 & 0.819 & 0.291\\
      $\text{ST}_\mathrm{l}$ & 0.232 & 0.841 & 0.304 \\
      $\text{MT}_\mathrm{l}$ & 0.236 & 0.804 & 0.288 \\\hline
      $\text{MGM}_\mathrm{/self}$ & 0.239 & 0.776 & 0.279 \\
      $\text{MGM}_\mathrm{/refine}$ & \bf 0.254 & 0.808 & 0.290 \\ 
      $\text{MGM}_\mathrm{recon}$ & 0.241 & 0.768 & 0.285 \\
      MGM & 0.251 &\bf 0.734 &\bf 0.273  \\ \hline
    \end{tabular}
    }
    \vspace{-10 pt}
    \caption{Ablation study. (1) $\text{ST}_\mathrm{l}$ and $\text{MT}_\mathrm{l}$: baselines with a larger number of parameters (with deeper backbones); (2) $\text{MGM}_\mathrm{/self}$: {\em without} self-supervision task; (3) $\text{MGM}_\mathrm{/refine}$: {\em without} classification refinement network; and (4) $\text{MGM}_\mathrm{recon}$: {\em with} a simple reconstruction task as self-supervision. The two proposed components are complementary and both benefit the multiple tasks. The refinement network works better for surface normal; the self-supervision network works better for semantic segmentation. Their combination achieves the best.}
    \label{tab:self}
    \vspace{-20 pt}
\end{table}

\begin{figure*}[t]
    \centering
    \begin{minipage}{0.27 \linewidth}
    \includegraphics[width =  \linewidth]{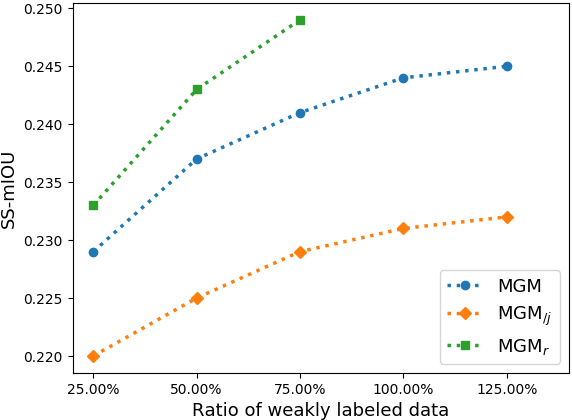}
    \end{minipage}
    \begin{minipage}{0.27 \linewidth}
    \includegraphics[width =  \linewidth]{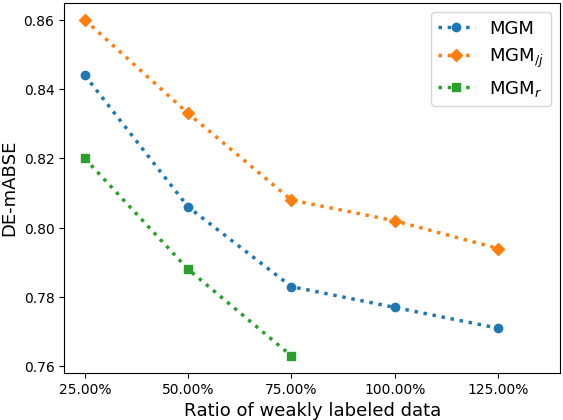}
    \end{minipage}
    \begin{minipage}{0.27 \linewidth}
    \includegraphics[width =  \linewidth]{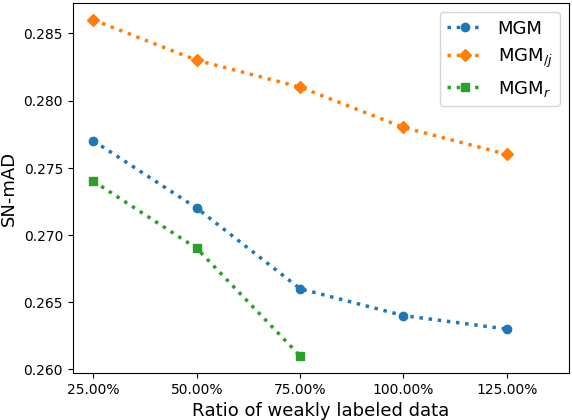}
    \end{minipage}
    \vspace{-10 pt}
    \caption{Performance change with different ratios of weakly labeled data. Joint learning significantly improves the performance. The performance of MGM keeps increasing with the number of the  weakly labeled {\em synthesized} images, achieving results almost comparable to that of $\text{MGM}_\mathrm{r}$ trained with all the available weakly labeled {\em real} images.}
    \label{fig:curve}
    \vspace{-14 pt}
\end{figure*}

\vspace{-2 pt}
\textbf{Number of Synthesized Images vs. Real images:} From the previous results, we have found that the synthesized images could benefit the target multi-tasks in a way similar to weakly labeled real images. To further investigate the impact of the number of synthesized images, we vary if from 25\% to 125\% during multi-task training on NYUv2 in the 25\% real data setting. Figure~\ref{fig:curve} summarizes the result. First, we can see that the performance gap between $\text{MGM}_\mathrm{/j}$ (without joint training) and MGM  becomes larger for a higher ratio of weakly labeled data, which indicates the importance of our joint learning mechanism. {\em More importantly}, while the real images are constrained in number due to the human collection effort, our generation network is able to synthesize \textit{unlimited} amounts of images. This is demonstrated in the comparison between $\text{MGM}_\mathrm{r}$ (with real images) and MGM: the performance of our MGM keeps improving with respect to the number of synthesized images, achieving results almost comparable to that of $\text{MGM}_\mathrm{r}$ when $\text{MGM}_\mathrm{r}$ uses all the available weakly labeled real images. 

We also provide ablation studies on the generalizability and the resolution of images in the supplementary material.

\vspace{-5 pt}
\subsection{Extension}

\textbf{Experiments with More Tasks:} MGM is also flexible and scalable with different tasks. In addition to the three tasks addressed in the main experiments, here we add three extra tasks: Edge Texture (ET), Reshading (Re), and Principal Curvature (PC), leading to six tasks in total. We evaluate the performance of all the compared models on Tiny-taskonomy in the 50\% data setting, and report the mean test loss for all the tasks. The result is reported in Table~\ref{tab:extension}. Again, our proposed method still outperforms state-of-the-art baselines.

\begin{table}[t]
    \centering
    \footnotesize
    \resizebox{\linewidth}{!}{
    \begin{tabular}{c|cccccc}
    \hline
    Model & SS ($\downarrow$) & DE ($\downarrow$) & SN ($\downarrow$) & ET ($\downarrow$) & Re ($\downarrow$) & PC ($\downarrow$) \\
      \hline
      ST &  0.120 & 1.768 & 0.157 & 0.228 & 0.703 & 0.462 \\
      MT & 0.112 & 1.747 & 0.169 & 0.241 & 0.704 & 0.436\\
      MGM & \bf 0.108 &\bf 1.715 &\bf 0.152 &\bf 0.201 &\bf 0.699 &\bf 0.417 \\ \hline 
    \end{tabular}
    }
    \vspace{-10 pt}
    \caption{Mean test losses for six tasks on Tiny-Taskonomy. Again, our MGM outperforms the baselines, indicating its flexibility, generability, and scalability.}
    \vspace{-20 pt}
    \label{tab:extension}
\end{table}

\vspace{-5 pt}
\section{Related Work}

\textbf{Multi-task Learning and Task Relationship:}
Multi-task learning (MTL) aims to leverage information coming from signals of related tasks so that each individual task can gain benefit~\cite{doersch2017multi}. \cite{ruder2017overview} identifies that most recent works use two clusters of strategies for MTL: hard parameter sharing techniques~\cite{kokkinos2017ubernet,doersch2017multi,pentina2017multi} and soft parameter sharing techniques~\cite{misra2016cross,sener2018multi,chen2018gradnorm}. These strategies have achieved good performance for MTL with similar tasks. Researchers have also carefully studied the task relationships among different tasks to make the best cooperations among them. \emph{Taskonomy} exploits the relationships among various visual tasks to benefit the transfer or multi-task learning~\cite{zamir2018taskonomy}. \cite{Pal_2019_CVPR} proposes a meta-learning algorithm to adapt existing models to zero-shot learning tasks. \cite{standley2019tasks} considers task cooperation and competition, and proposes a method to assign tasks to a few neural networks to balance all of them. Some other following works also explores task relationships among different types of tasks~\cite{sun2019adashare,zamir2020robust,armeni20193d}. These works only consider MTL with discriminative tasks. In comparison, we first introduce generative modeling to multi-task visual learning.

\textbf{Generative Modeling for Visual Learning:}
Besides the initial goal of synthesizing realistic images, some recent work has explored the potential to leverage generative models to synthesize ``usefull'' images for other visual tasks~\cite{shorten2019survey}. The most straightforward way is to generate images and the corresponding annotations as data augmentation for the target visual task~\cite{bao2020bowtie,sandfort2019data, choi2019self}. Besides, \cite{wang2018low} proposes to generate imaginary latent features rather than images to better benefit the low-shot classification. Another strategy to leverage generative models is through well-designed error feedback or adversarial training~\cite{luc2016semantic,cs2018monocular,mustikovela2020self}. There have been works that apply generative models for different visual tasks including classification~\cite{zhan2018verisimilar,frid2018gan,zhu2018emotion}, semantic segmentation~\cite{souly2017semi,luc2016semantic} and depth estimation~\cite{pilzer2018unsupervised,Aleotti_2018_ECCV_Workshops}. These methods are limited to a single specific task and have relatively low generalizability for more tasks. In comparison, MGM is applicable to various multiple visual tasks and different generative networks. 

\textbf{Reduced-Supervision Methods:}
Recent works take advantage of weakly labeled data by assigning some self-created labels (\eg colorization, rotation, reconstruction)~\cite{noroozi2016unsupervised,noroozi2017representation,chen2020simple,dosovitskiy2014discriminative,pathak2016context}. Similar self-supervised techniques have been proved useful for multi-task learning~\cite{liu2008semi,ren2018cross,doersch2017multi,lee2019multi}. Among these techniques, a famous one is the \textit{Expectation-Maximization (EM)} algorithm~\cite{dempster1977maximum}, which leverages the information of weakly or unlabelled data by iteratively estimating and refining their labels. \cite{papandreou2015weakly} further applies \textit{EM} algorithm for semi-supervised semantic segmentation. We adopt a similar spirit and introduce the refinement network for MGM framework.

\vspace{-5 pt}
\section{Conclusion}

Motivated by multi-task learning of shareable feature representations, this paper proposes to introduce generative modeling for multi-task visual learning. The main challenge is that it is hard for generative models to synthesize both RGB images and pixel-level annotations in multi-task scenarios. We address this problem by proposing multi-task oriented generative modeling (MGM) framework equipped with the self-supervision network and the refinement network, which enable us to take advantage of synthesized images paired with image-level scene labels to facilitate multiple visual tasks. Experimental results indicate our MGM model consistently outperforms state-of-the-art multi-task approaches.

{\small
\bibliographystyle{ieee_fullname}
\bibliography{egbib}
}

\end{document}